\documentclass[preprint]{article}

\usepackage{neurips_2023}

\usepackage[utf8]{inputenc} % allow utf-8 input
\usepackage[T1]{fontenc}    % use 8-bit T1 fonts
\usepackage{hyperref}       % hyperlinks
\usepackage{url}            % simple URL typesetting
\usepackage{booktabs}       % professional-quality tables
\usepackage{amsfonts}       % blackboard math symbols
\usepackage{nicefrac}       % compact symbols for 1/2, etc.
\usepackage{microtype}      % microtypography
\usepackage{xcolor}         % colors
\usepackage{graphicx}
\usepackage{tabularx}
\usepackage{multirow}
\usepackage{color}
\usepackage{colortbl} 
\usepackage{soul}
\usepackage{svg}

\usepackage{natbib}
\setcitestyle{numbers,square}

\title{Towards Total Online Unsupervised Anomaly Detection and Localization in Industrial Vision}

\author{ 
    Han Gao$^{1,2}$\thanks{These authors contributed equally to this work.} \\
    \texttt{gaohan2022@ia.ac.cn} \\
    \And
    Huiyuan Luo$^{1,2}$\footnotemark[1] \\
    \texttt{huiyuan.luo@ia.ac.cn} \\
    \And
    Fei Shen$^{1,2,3}$ \\
    \texttt{fei.shen@ia.ac.cn} \\
    \And
    Zhengtao Zhang$^{1,2,3,4}$\thanks{Corresponding author.}\\
    \texttt{zhengtao.zhang@ia.ac.cn} \\
}

\begin{document}

\maketitle

\footnotetext[1]{Engineering Laboratory for Intelligent Industrial Vision, Institute of Automation, Chinese Academy of Sciences, Beijing 100190, People’s Republic of China}
\footnotetext[2]{University of Chinese Academy of Sciences, Beijing 100049, People’s Republic of China}
\footnotetext[3]{CASI Vision Technology CO., LTD., Luoyang 471000, People’s Republic of China}
\footnotetext[4]{Binzhou Institute of Technology, Weigiao-UCAS Science and Technology Park, Binzhou, Shandong 256606, People’s Republic of China}

\begin{abstract}
 Although existing image anomaly detection methods yield impressive results, they are mostly an offline learning paradigm that requires excessive data pre-collection, limiting their adaptability in industrial scenarios with online streaming data. Online learning-based image anomaly detection methods are more compatible with industrial online streaming data but are rarely noticed. For the first time, this paper presents a fully online learning image anomaly detection method, namely \textit{LeMO}, learning memory for online image anomaly detection. LeMO leverages learnable memory initialized with orthogonal random noise, eliminating the need for excessive data in memory initialization and circumventing the inefficiencies of offline data collection. Moreover, a contrastive learning-based loss function for anomaly detection is designed to enable online joint optimization of memory and image target-oriented features. The presented method is simple and highly effective. Extensive experiments demonstrate the superior performance of LeMO in the online setting. Additionally, in the offline setting, LeMO is also competitive with the current state-of-the-art methods and achieves excellent performance in few-shot scenarios. 
 % Source code will be released upon paper acceptance.
\end{abstract}

\section{Introduction}

Anomaly detection, which capitalizes on normal samples to identify outliers without the need for laborious sample annotation, is particularly effective in fields like industrial vision defect detection, where acquiring labeled samples is challenging. Recently, deep learning-based methods take unsupervised image anomaly detection task as a problem of out-of-distribution detection\cite{patchcore,you2022unified,wyatt2022anoddpm,li2021cutpaste,jiang2022softpatch,xie2023pushing,batzner2023efficientad}. These methods typically follow two main phases: data collection and offline model training. Initially, a high-quality, diverse dataset is constructed, large enough to accurately represent the normal samples' distribution. Subsequently, model training is conducted, repeatedly using the collected samples to fit the true distribution of normal patterns. Therefore, these methods can be considered as offline learning-based anomaly detection methods. Despite promising results, such methods are primarily suitable for the detection scenes with low immediacy requirements.

% Anomaly detection, which capitalizes on normal samples to identify outliers without the need for laborious sample annotation, is particularly effective in fields like industrial vision defect detection, where acquiring labeled samples is challenging. Recently, deep learning-based methods take unsupervised image anomaly detection task as a problem of out-of-distribution detection\cite{patchcore,li2021cutpaste,jiang2022softpatch,cfa,spade,padim,f:divide,rippel2021modeling,bergmann2020uninformed,salehi2021multiresolution,deng2022anomaly,wang2021student,f:cflow,fastflow,rudolph2021same,ruff2018deep,panda,Patchsvdd,zaheer2020old,sabokrou2018adversarially,you2022unified,hou2021divide,draem,yan2021learning,zhou2020encoding,shi2021unsupervised,liu2020towards,akcay2019ganomaly,bergmann2018improving}. These methods typically follow two main phases: data collection and offline model training. Initially, a high-quality, diverse dataset is constructed, large enough to accurately represent the normal samples' distribution. Subsequently, model training is conducted, repeatedly using the collected samples to fit the true distribution of normal patterns. Therefore, these methods can be considered as offline learning-based anomaly detection methods. While they have yielded promising results, they are primarily suitable for the detection scenes with low immediacy requirements.

However, offline learning-based anomaly detection algorithms encounter several problems when applied in real industrial  scenarios. These manifest as follows: firstly, with the rise of customized flexible production modes, rapid changes in industrial production lines are commonplace. But existing offline anomaly detection methods typically require extensive collection and storage of normal samples, which can consume substantial memory and time resources, delayed to meet the rapid switching needs of production line models. Secondly, the performance of the algorithm is heavily reliant on the size of the normal samples dataset, but it is difficult to assess how many normal samples are actually needed. At the same time, in pursuit of high performance, existing algorithms often equip with heavy parameters and complex model structures, leading to longer overall training times and the consumption of considerable computational resources, which are usually scarce in industrial scenarios. Lastly, the industrial environment is more complex. Interference factors such as noise and unstable image acquisition equipment may lead to the occurrences of image feature drift. Existing algorithms, which are trained solely based on collected data, cannot swiftly recognize changes in new data, leading to diminished performance over time. These issues pose significant barriers to the application and deployment of existing offline anomaly detection algorithms, and consequently limiting the practical value of such anomaly detection algorithms.

\begin{figure}[ht]
 \label{fig:online}
  \centering
   \includegraphics[width=\textwidth]{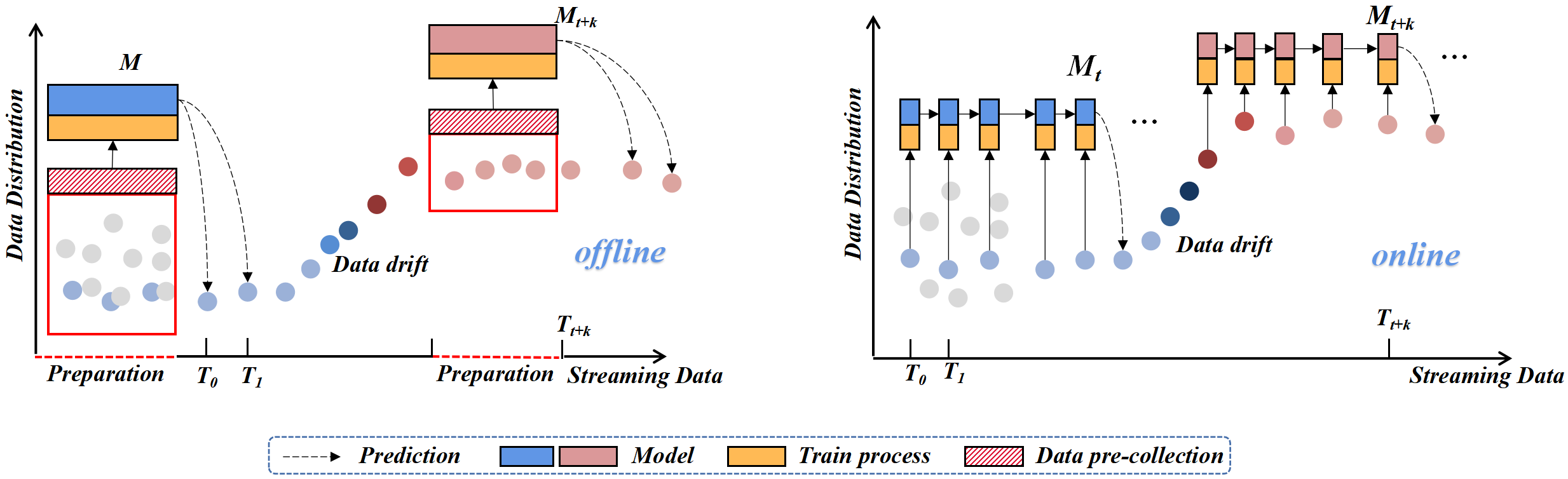}
  \caption{Offline anomaly detection can only be trained repeatedly by pre-collecting a large amount of data over a certain period. In contrast, online anomaly detection accepts input from online streaming data, eliminating the need for data pre-collection and allowing the model to swiftly adapt to data drift.}
\end{figure}

% Fundamentally, the cause of the above problems is that existing algorithms are static and offline, while data is dynamic and online. To address these issues, in this paper, we propose a new paradigm for anomaly detection:\textit{ normal samples are not collected in advance but input as a data stream.} The model only inputs one image at a time and conducts lightweight, rapid online training to possess good detection capabilities. We define this new learning-based paradigm as an online learning image anomaly detection algorithm. A visual representation of two learning paradigms is shown in Fig.\ref{fig:online}. In the online learning paradigm, samples should be input in a stream-like manner to avoid extensive sample collection and storage processes, thus saving considerable memory and time. Additionally, it should generally be lightweight, requiring no lengthy training time, significantly shortening the algorithm deployment cycle. Furthermore, online learning anomaly detection algorithms should learn from online feedback data, adjust the model in real-time, respond immediately to online changes, and exhibit superior environmental adaptability and robustness. However, to the best of our knowledge, online learning-based image anomaly detection algorithms have yet to attract substantial attention and proposal.

Fundamentally, the cause of the above problems is that existing algorithms are static and offline, while data is dynamic and online. To address this challenge, we propose a new paradigm for anomaly detection:\textit{ normal samples are not collected in advance but input as a data stream.} The model processes only one image at a time and conducts swift and lightweight online training to achieve good detection capabilities. We define this new learning-based paradigm as an online learning image anomaly detection algorithm, visually depicted in Fig.\ref{fig:online}. Under the online learning paradigm, samples should be input in a stream-like manner to avoid extensive sample collection and storage processes, thus saving considerable memory and time. Furthermore, it should generally be lightweight, avoiding long training periods and significantly reducing the algorithm deployment cycle. Online learning anomaly detection algorithms should learn from online feedback data, adjusting the model in real-time to swiftly respond to online changes, thereby demonstrating superior adaptability and robustness to environmental changes. However, to the best of our knowledge, online learning-based image anomaly detection algorithms have yet to attract substantial attention and proposal.

In this paper, we first define and propose a fully online image anomaly detection method, \textit{LeMO}. Given that feature embedding-based memory methods offer a simple model structure and great scalability, we also adopts a feature embedding-based memory architecture. However, unlike previous feature embedding methods (\textit{i.e.}PatchCore \cite{patchcore} and CFA\cite{cfa}) that require an initialization step for the memory bank, LeMO directly uses orthogonally randomized noise to initialize and fill the prototype memory bank. Concurrently, we design a contrastive learning loss function to jointly optimize the memory and the target-oriented features of the objective category. This design allows LeMO to achieve fully online training and inference. Numerous experiments demonstrate that LeMO achieves the best detection performance in an online setting. Additionally, even in offline scenarios, LeMO performs on par with the state-of-the-art methods.
%In summary, our contributions are:
%\begin{itemize} 
%\item[$\bullet$] To the best of our knowledge, we are the first to focus on image anomaly detection algorithms under the online learning paradigm. It can better meet the detection needs in industrial scenarios, especially in assembly line manufacturing scenarios, and possesses superior practicality and environmental adaptability. \item[$\bullet$] We propose a total online learning-based image anomaly detection method. The method only uses orthogonal randomized noise to initialize memory, eliminating the process of extensive data collection. A contrastive learning loss function is proposed to jointly optimize memory with the image's target-oriented features. 
%\item[$\bullet$] In an online learning setting, our proposed method achieves image anomaly classification accuracy of 96.7\%, 86.6\%, and 94.3\% on the MVTec AD, MPDD, and VisA datasets, respectively. In an offline learning setting, our method's accuracy is competitive with the current SOTA method and achieves excellent performance in the few-shot scenarios. \end{itemize}

\section{Related works}
\subsection{Feature embedding}

Feature embedding methods can be broadly classified into four categories: memory bank\cite{cfa,patchcore,spade,padim,f:divide,rippel2021modeling,jiang2022softpatch}, knowledge distillation\cite{bergmann2020uninformed,salehi2021multiresolution,deng2022anomaly,wang2021student}, normalizing flow networks\cite{f:cflow,fastflow,rudolph2021same}, and one-class classification\cite{ruff2018deep,panda,Patchsvdd,zaheer2020old,sabokrou2018adversarially}. Among these, the memory-based approach is the most straightforward and efficient, with representative methods including SPADE\cite{spade}, Padim\cite{padim}, PatchCore\cite{patchcore}, and CFA\cite{cfa}. These methods operate in such a pipeline: during the training phase, the feature extraction backbone extracts features from all normal images and stores them in a memory bank. The testing phase can be viewed as a cross-attention process, where the test sample's features serve as a query to match the key of normal features stored in the memory. To enhance the representation of normal pattern and reduce the size of memory, SPADE\cite{spade} uses a multi-resolution pyramid structure. Padim\cite{padim} reduces the size of the memory bank through a probabilistic estimation method. PatchCore\cite{patchcore} only uses intermediate layer features to reduce the feature bias and apply a coreset algorithm to sample the memory, and CFA\cite{cfa} proposes a coupled-hypersphere-based feature adaptation method to obtain target-oriented features and designs a compression scheme to reduce the memory's size. However, these methods face two main issues: the performance heavily relies on the completeness of the memory, requiring extensive collection of normal images for a comprehensive representation of normal pattern; and the memory size is often related to the dataset size or image size, making them unsuitable for large data or large size image scenarios. These limitations significantly restrict the practicality of memory-based methods in real-world scenarios.

\subsection{Reconstruction-based anomaly detection}
% Reconstruction-based methods consider anomaly detection at the image dimension. Some uses auto-encoders or generators to encode and decode the input normal images, indirectly learning the distribution of normal images by reconstructing the input image. Such algorithms are intuitive and explainable, but their generation effects are relatively poor for textures or objects with complex structures. Furthermore,they can easily lead to an identity mapping situation where anomalies are also well reconstructed. Recently, DRAEM has constructed defect pseudo-labels by simulating defects and used a reconstruction network to restore the pseudo-labels, and then used a discriminative segmentation network to detect and localize the anomaly. Reconstruction-based methods are one of the mainstream methods in anomaly detection, but they inherently have generalization and reconstruction error problems, and their network structures are generally more complex, running slower, making them unsuitable for use in industrial scenarios where timeliness is crucial.

Reconstruction-based methods\cite{you2022unified,hou2021divide,draem,yan2021learning,zhou2020encoding,shi2021unsupervised,liu2020towards,akcay2019ganomaly,bergmann2018improving} tackle anomaly detection at the pixel level. Some uses auto-encoders\cite{shi2021unsupervised,liu2020towards,bergmann2018improving,zhou2020encoding} or generators\cite{akcay2019ganomaly,yan2021learning,draem,hou2021divide} to encode and decode the input normal images, indirectly learning the normal image distribution through reconstruction. While these algorithms offer intuitive and explainable results, they often struggle with textures or objects of complex structures. Moreover, they might lead to an identity mapping situation where anomalies are reconstructed as well. Recently, DRAEM\cite{draem} has constructed defect pseudo-labels by simulating defects and used a reconstruction network to restore the pseudo-labels, and then used a discriminative segmentation network to detect and localize the anomaly. UniAD\cite{you2022unified} tackles this issue with a layered query decoder, a neighbor masked attention module, and a feature jittering strategy. Though reconstruction-based methods are prevalent in anomaly detection, they inherently suffer from generalization and reconstruction error issues. Moreover, their network structures generally tend to be more complex and run slower, making them less suitable for time-sensitive industrial scenarios.

\subsection{Anomaly detection in streaming data}
Anomaly detection in streaming data is mostly applied in the fields of sensor data\cite{Sensor}, logs\cite{logs}, and other time series or high-dimensional data\cite{TimeSeries,TimeSeAD}, where the task is mostly to identify abnormal events or behaviors from normal time series. According to \cite{TimeSeAD}, there are two mainstream methods for AD on time series: detect anomalies at each time step separately (point-wise) or the entire time series (globally). Point-wise methods are naturally suited for real-time prediction and anomaly localization, which is very meaningful for practical scenarios. Among them, transformer-based methods\cite{TransAD,TransAnom} have achieved excellent performance due to their superior long-sequence modeling ability. But these algorithms are not directly applicable to online image anomaly detection scenarios for industrial vision. This is because industrial images are more complex but relatively stable in semantic content, eliminating the need for long-sequence time modeling capabilities.

\section{The proposed method}
\subsection{Overview}
%Previous deep feature embedding-based approaches necessitate the storage of a comprehensive dataset memory bank or alignment of image features to prototypes in the memory bank, which prevents their application for online anomaly detection scenarios. In fact, we explore that memory bank modeling and feature updating can be done uniformly in a learnable architecture. The model exhibits adaptability, flexibly adjusting to the drift in input data. When data features shift, we dynamically update the prototype memory bank $\mathcal{P} $ in real-time to adapt to changes in data distribution.

\begin{figure}[ht]
  \centering
   \includegraphics[width=0.85\textwidth]{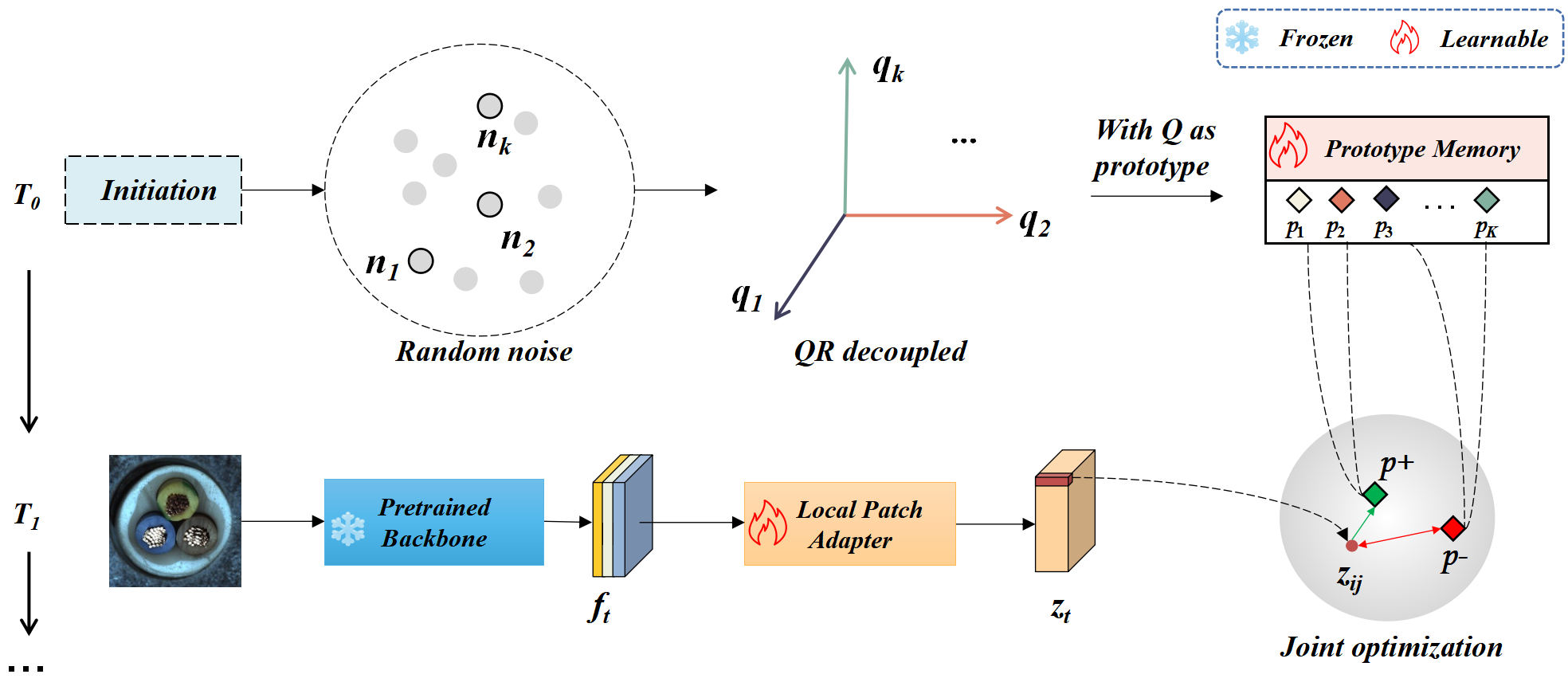}
  \caption{Overview of the proposed method.}
  \label{fig:method}
\end{figure}

In this paper, we propose a unified online anomaly detection architecture to jointly update the prototype memory bank $\mathcal{P}$ and pre-trained image features $\mathcal{F}$, which can capture the most discriminative positive sample patterns and eliminate the pre-trained feature bias simultaneously. Formally, as depicted in Fig.\ref{fig:method}, our proposed approach consists of an encoder, a local patch adapter, and a prototype memory bank $\mathcal{P} =(p_1,...,p_{K})$. During the training phase, the encoder maps the input streaming data $x_t $  at time $t$ into lower-resolution multi-scale visual representations ${f_{t}}$ . Subsequently, the ${f_t}$ representations are concatenated and augmented with coordinate position information by the local patch adapter to enhance spatial awareness capabilities. The enhanced features are then passed through a non-linear projector, which is implemented as a 1x1 convolutional layer with learnable parameters (See Appendix A for details). This process yields locally enhanced learnable features $ \mathit{z}_ {t} \in \mathbb{R}^{D \times H \times W}$. It's worth noting that the encoder is pre-trained on ImageNet and will remain frozen. $D$, $H$ and $W$ are the feature dimension, height and width of $ \mathit{z}_ {t} $, respectively. $K$ is the number of prototypes.

During the training period $ 0\rightarrow{t}$, the contrastive anomaly detection loss function AnoNCE is optimized. During this process, the feature $\mathit{z}_{ij}=\mathit{z_{t}}(i,j)$, $i\in [1,...,H]$ and $j\in [1,...W]$ is pulled towards the most relevant prototypes $p^+$ and pushed away from unrelated prototypes $p^-$. Concurrently, the memory bank $\mathcal{P}$ undergoes an update. At the inference stage at $t+1$, it is assumed that the model has developed the ability to accurately map $z(i,j)$ from normal samples to the most related prototype vector $p^*$. The prototype memory bank at this time is represented as $\mathcal{P}^{*}$. Anomaly detection and localization are performed based on the degree of match between the feature $z_{t+1}(i,j)$ of the test sample and $\mathcal{P}^{*}$.

% During the training period $ 0\rightarrow{t}$, the contrastive anomaly detection loss function AnoNCE is optimized. During this process, the feature $ z_{ij}=z_{t}(i,j)$, $i\in [1,...,H]$ and $j\in [1,...W]$ is pulled towards the most relevant prototypes $p^+$ and pushed away from unrelated prototypes $p^-$. Concurrently, the memory bank $\mathcal{P}$ undergoes an update. At the inference stage at $t+1$, it is assumed that the model has developed the ability to accurately map $z_{ij}$ from normal samples to the most related prototype vector $p^*$. The prototype memory bank at this time is represented as $\mathcal{P}^{*}_{t} $. Anomaly detection and localization are performed based on the degree of match between the feature $z_{t+1} $of the test sample and $\mathcal{P}^{*}_{t} $.

Our proposed approach exhibits significant flexibility, allowing for numerous variants at each stage in pursuit of optimal training/inference speed or improved anomaly detection accuracy within an online environment. Further specific discussions are offered in next sections, including the initialization strategy for memory bank (section\ref{sec:init_memory}), the online joint optimization method for image features and memory bank (section\ref{sec:online_update}), and the computation of anomaly scores (section\ref{sec:cal_score}).

\subsection{Initialization of prototype memory bank}
\label{sec:init_memory}
We assume the memory-based approach is an essentially a maximally equivalent replacement of the high-dimensional feature space with a representative and discriminative low-dimensional subspace, \textit{i.e.} cluster centroid. Previous methods use various sample techniques, such as coreset\cite{patchcore}, $\mathit{k}$-means\cite{cfa}, patch distribution modeling\cite{padim}. However, the inherent flaw is that they all first need a complete repository to characterize the normal pattern. But it is theoretically agnostic to know how many the normal samples are adequate, as well, preparing enough data is tough and inefficient. To track such issues, we propose that the basis vectors may be an alternative technique to find the optimal equivalent replacement, which eliminates the data pre-collection process. Each basis vector can be considered as a prototype for the local fine-gained normal pattern. In this paper, we assemble $\mathit{K}$ basis vectors as a prototype memory bank $\mathcal{P}$. 

Our goal is to jointly update the prototype memory bank $\mathcal{P}$ and target-oriented image features $ \mathit{z} $ in an online anomaly detection scenario. As such, an initialization of $ \mathcal{P}$ is needed at time $t=0$. We propose two efficient initialization strategies: 1) based on a random single image, and 2) based on decoupled random noise. These strategies obviate the need for extensive normal sample data collection, requiring minimal resources to effectively initialize the memory bank. We especially advocate for initializing $\mathcal{P}$ using decoupled random noise, as it avoids the use of any normal data as prior information (which is unavailable at time $t=0$), facilitating an undisturbed and swift commencement of online model training.

\subsubsection{Random single image initiation}
Based on the above considerations, the first method of initializing the prototype memory bank in this paper is to randomly select a normal sample at the start of training. Visual features $ \mathit{z}_ {0} \in \mathbb{R}^{D \times H \times W}$ are obtained through the encoder and patch adapter, and then are performed by a standard clustering method (such as $\mathit{k}$-means) to fill with $\mathit{K}$ prototypes. This provides strong diversity and normal sample prior information for the initialization phase, allowing the model to dynamically optimize the prototype memory bank and the paramters of local path adapter during the online training process.

\subsubsection{Decoupled noise initiation}
% Although the strategy of initialization based on random images is simple and efficient, it still introduces prior information from the normal sample. To pursue total online learning anomaly detection, we propose a method that initializes the memory bank using orthogonal random noises, which can simulates the feature deconstruction process with images effectively. Under the online learning setting, random noise ensures our model can start online quickly without disturbance, while the orthogonalization step guarantees the decoupling nature of the initialization filling vectors, providing guidance for the update of the memory bank and avoiding the possibilities of falling into locally optimal solutions. As shown in the table, our initialization approach achieves total online anomaly detection with the least complexity and memory bank size. Furthermore, the number of prototypes $K$ can be chosen arbitrarily (not less than 1 for AnoNCE loss function in section \ref{sec:3.3.1-loss}, we choose $K=10$ in this paper). Note that the computational complexity and the size of memory bank in other methods correlate with the dataset scale or the image size, whereas our method can set $K$ independently of these factors. This demonstrates the flexibility and high efficiency of our method during training and inference.

Though simple and efficient, initializing with random images introduces prior normal sample information. To enable total online learning anomaly detection, we propose to initialize the memory bank with orthogonal random noises, effectively simulating the feature deconstruction process. This allows a quick start without disturbance, and the orthogonalization step ensures the decoupling of initialization vectors, guiding memory bank updates and avoiding local optimal solution. In this paper, we apply QR decomposition to orthogonalize the random noises. QR decomposition is an algorithm that decomposes a matrix $\mathit{A}$ into a matrix $\mathit{Q}$ with standard orthogonal column vectors and an upper triangular matrix $\mathit{R}$. $\mathit{Q}$ is denoted as the initial orthogonal prototypes memory bank $\mathcal{P} $, which makes a well foundation for the subsequent learnable optimization.

Spatial complexity and memory size of the memory bank construction process of previous anomaly detection methods and our method's initialization are analyzed in Table \ref{memory bank size}. $\mathcal{X}$ denotes the dataset scale and $\gamma$ is the compression ratio, \textit{i.e.}, the number of cluster centers. $H,W,D$ are the dimensions of the feature $ \mathit{z} $. Our approach achieves fully online anomaly detection with the least complexity and memory bank size. The number of prototypes $K$ is flexible. We have chosen $K=10$ in this paper. Unlike other methods, the spatial complexity and memory bank size of our method do not depend on the scale of the dataset or the size of images, which highlights the flexibility and efficiency.

\begin{table}[ht]\scriptsize
\caption{Analysis of memory bank modeling complexity and memory bank size.}
\label{memory bank size}
\centering
\begin{tabular}{l|c|c|c|c|c}
\toprule
Methods & SPADE\cite{spade} & PaDiM\cite{padim} & PathCore\cite{patchcore} & CFA\cite{cfa} & Ours \\
\midrule
Modeling & $\mathcal{O}(|\mathcal{X}|HWD)$ & $\mathcal{O}(|\mathcal{X}|HWD^2)$ & $\mathcal{O}(|\mathcal{X}|HWD)$ & $\mathcal{O}(HWD)$ & $\mathcal{O}(KD)$ \\
\midrule
Memory bank & $\mathbb{R}^{|\mathcal{X}|\times H\times W\times D}$ & $\mathbb{R}^{H\times W\times D^2}$ & $\mathbb{R}^{|\mathcal{X}|\times\gamma(H\times W)\times D}$ & $\mathbb{R}^{\gamma(H\times W\times D)}$ & $\mathbb{R}^{K\times D}$ \\
% \midrule
% Modeling & $\mathcal{O}(|\mathcal{X}|HWD)$ & $\mathcal{O}(|\mathcal{X}|HWD^2)$ & $\mathcal{O}(|\mathcal{X}|HWD)$ & $\mathcal{O}(HWD)$ & $\mathcal{O}(KD)$ \\
% \midrule
% Memory Bank & $\mathcal{G}\in\mathbb{R}^{|\mathcal{X}|\times H\times W\times D}$ & $\mathcal{N}(\mu,\Sigma)\in\mathbb{R}^{H\times W\times D^2}$ & $\mathcal{M}\in\mathbb{R}^{|\mathcal{X}|\times\gamma(H\times W)\times D}$ & $\mathcal{C}\in\mathbb{R}^{\gamma(H\times W\times D)}$ & $\mathcal{P}\in\mathbb{R}^{K\times D}$ \\
\bottomrule
\end{tabular}
\end{table}

\subsection{ Online joint optimization}
\label{sec:online_update}
\subsubsection{Loss function}
\label{sec:3.3.1-loss}
In the training phase depicted in Fig.\ref{fig:method}, we assign feature $\mathit{z}_{ij}$ at time $t$ to the most relevant prototype within the memory bank $\mathcal{P} =(p_1,...,p_{K}) \in \mathbb{R}^{K \times D}$, and optimize the loss to enhance their similarity. Given that industrial images often contain complicated objects or uniform textures, feature $\mathit{z}_{ij}$ could potentially align with multiple prototypes. Consequently, we design a contrastive over-cluster anomaly detection loss based on NCE\cite{gutmann2010noise}, which we refer to as AnoNCE:
\begin{equation}\footnotesize
\label{eqa:loss}
\mathit{L}_{z_{ij},\left\{p^{+}\right\},\left\{p^{-}\right\}}=-\log \frac{\sum_{p^{+}} \exp \left({sim}\left(\mathit{z_{ij}}, \mathit{p}^{+}\right) / \tau\right)}{\sum_{p^{+}} \exp \left({sim}\left(\mathit{z}_{ij}, \mathit{p}^{+}\right) / \tau\right)+\sum_{p^{-}} \exp \left({sim}\left(\mathit{z}_{ij}, \mathit{p}^{-}\right) / \tau\right)}
\end{equation}

We view each $\mathit{z}_{ij}$ as an anchor, the similar prototypes as positive samples, and the dissimilar ones as negative samples. During the training phase, $z_{ij}$ will be updated to gather around the positive samples while moving away from the negative samples. The similarity $sim(\mathit{z}_{ij}, \mathit{p}) $ is a measure of the similarity between the sample feature $\mathit{z}_{ij}$ and the prototype vector $\mathit{p}$. We choose a sharp temperature coefficient $ \tau$, which implies a higher degree of attention to hard negative samples.

\subsubsection{Learning-based memory update}
\label{learning based strategy}
While minimizing the loss function \ref{eqa:loss}, the prototypes in $\mathcal{P}$ will also be jointly optimized during the training process. There are multiple choices for the similarity function $sim(z,p)$. In this paper, we choose the Euclidean distance due to its higher sensitivity to anomalies, which benefits detection accuracy. The ultimate optimization goal of our method is:

\begin{equation}\footnotesize
\min_{z,p} E(z, p)=-\frac{1}{HW} \sum_z \log \frac{\sum_{\mathit{p}^{+}} \exp \left(-m\left(\left|z_{ij}-\mathit{p}^{+}\right|, r\right) / \tau\right)}{\sum_{\mathit{p}^{+}} \exp \left(-m\left(\left|z_{ij}-\mathit{p}^{+}\right|, r\right) / \tau\right)+\sum_{\mathit{p}^{-}} \exp \left(-m\left(\left|z_{ij}-\mathit{p}^{-}\right|, r\right) / \tau\right)}
\end{equation}

The function $m(d,r)$ equals $max(d-r,0)$, which relaxes the similarity constraint by a margin $r$ for enhanced robustness. Our method aims to obtain an optimal set of prototypes that accurately represent the true distribution of normal samples, aligning with the goal of maximizing the mutual information between local image features and corresponding prototypes\cite{oord2018representation}.

\subsubsection{Feature enhanced memory update}
\label{feature based strategy}
% Despite the excellent detection performance achieved by optimizing $ E(z,p)$ using the learning-based memory bank update strategy, there is still a risk of degenerate solutions that could potentially map the majority of an image's features to a single prototype. Ideally, all of an image's features ${z}$ should be uniformly allocated across multiple prototypes, thus forming several stable hyperspheres to circumvent degenerate solutions and improve accuracy. Therefore, after updating the ${z}$ within an image, we modify the results of label allocation when aligning ${z}$ to prototypes, ensuring an approximately equal number of features corresponding to each prototype vector. In this strategy, prototypes are no longer jointly updated but are updated as centroids of all features with identical labels in the newly allocated labels. Specifically, if a prototype's corresponding group has fewer features than a designated threshold, we merge it with the group possessing the most features, execute a $\mathit{k}$-means operation, and split it into two new groups, selecting the smaller one to replace the original small group. Although such technique has been mentioned in Online DeepCluster, we apply it at a fine grained feature level and are the first to demonstrate its beneficial effect in anomaly detection tasks.

Despite the excellent detection achieved by optimizing $E(z,p)$ via the learning-based memory bank update strategy, there's a risk of degenerate solutions. Ideally, each image's features ${z}$, should be uniformly distributed across prototypes, forming stable hyperspheres to avoid degenerate solutions and enhance accuracy. Therefore, after updating the ${z}$ in an image, we adjust the label assignments when aligning ${z}$ to prototypes, so as to balance the number of features corresponding to each prototype vector. In this approach, prototypes are updated as centroids of all features with identical labels in the new allocation. If a prototype's group has fewer features than we expect, we merge it with the largest group, perform $\mathit{k}$-means operation, and reformulate two new groups. Although this technique has been referenced in \cite{odc}, we are the first to apply it at a fine-grained feature level and demonstrate its efficacy in anomaly detection tasks. More details could be found in Appendix A.

\subsection{Anomaly score}
\label{sec:cal_score}
We follow the standard procedure of previous feature embedding-based anomaly detection methods to compute anomaly scores. For the test features $\mathit{z}_{test}\in \mathbb{R}^{D \times H \times W}$, we calculate the degree of matching with the updated memory bank $\mathcal{P}^{*}$ as $\mathcal{S}$:
\begin{equation}\footnotesize
\mathcal{S}(i, j)=\min _k\left|z_{test}(i,j)-p^{*}_k\right|
\end{equation}

The anomaly score $\mathcal{A}$ can be formulated as following:
\begin{equation}\footnotesize
\mathcal{A}(i, j)=\frac{\exp (-\mathcal{S}(i, j))}{\sum_{p^{*}} \exp (-|z_{test}-p^{*}|)} \cdot \mathcal{S}(i, j)
\end{equation}

\section{Experiments}
\subsection{Experimental details}

Our method is evaluated on the VisA\cite{visa}, MVTec AD\cite{mvtec}, and MPDD\cite{mpdd} benchmark datasets, employing Image-level AUROC (I-AUROC), Pixel-level AUROC (P-AUROC), and Pixel-level AUROC (P-AUROC) metrics for anomaly detection and location. We compare our method with four SOTA anomaly detection methods: PatchCore\cite{patchcore}, CFA\cite{cfa}, FastFlow\cite{fastflow}, and DRAEM\cite{draem}. The evaluation of these methods is performed using anomalib\cite{anomalib}. Our method adopts the Adam\cite{kingma2014adam} optimizer with a base learning rate of 1e-3 and weight decay of 5e-4. The hyperparameter $ \mathit{r} $ in AnoNCE is 1e-5. A separate model is trained for each class. The feature extractor backbone is the Wide-ResNet50\cite{wideresnet} model. Images are initially resized to $256 \times 256$, then centrally cropped to $224 \times 224$ and normalized. During feature extraction, we combine layer2, layer3, layer4 of the pre-trained Wide-ResNet50 backbone, add coordinate information, and apply a 1x1 convolutional layer to yield the enhanced features. The prototype memory bank $\mathcal{P}$ contains $K=10$ prototype vectors. All methods are implemented in PyTorch and utilize a single NVIDIA 3090 GPU.

\subsection{Online learning}
In this section, we train the model for one epoch with a batch size of 1 to simulate the handling of online streaming data. Moreover, since most existing anomaly detection methods are designed for offline scenarios, we modify them to suit online scenarios, ensuring a comprehensive and fair comparison with our method. Specifically, we incorporate the exponential moving average (EMA) into the updating strategy of PatchCore\cite{patchcore}'s coreset construction and CFA\cite{cfa}'s feature bank construction, enabling these methods to dynamically update the features stored in their memory banks with online stream data input. The anomaly detection and localization performance of our method is evaluated in terms of efficiency and adaptability as follows. More details are in Appendix A.

\subsubsection{Efficient online anomaly detection}
\label{exp_sec:efficiency}
\textbf{Accuracy report.} Table \ref{online performance comparison} presents the I-AUROC, P-AUROC, and P-AUPRO results on the VisA\cite{visa}, MVTec AD\cite{mvtec}, and MPDD\cite{mpdd} datasets. Although existing SOTA anomaly detection methods are adapted to the online setting, our method still outperforms them. For the image-level evaluation metric I-AUROC, our method surpasses the highest accuracy of other methods by 1.9\% and 0.6\% on the VisA and MVTec AD datasets, respectively. For the pixel-level anomaly localization evaluation metric P-AUROC, our method achieves the highest accuracy on all three datasets. Additionally, although our method lags behind PatchCore\cite{patchcore} by a maximum of 3.4\% on the P-AUPRO metric, PatchCore\cite{patchcore} shows substantial fluctuations in other metrics, for example, a notable difference of 14.2\%, 10.0\%, and 15.3\% on the I-AUROC metric across the three datasets, respectively. This demonstrates that our method exhibits more stable and superior performance in the online setting.

\begin{table}[h]\scriptsize
\caption{Online accuracy comparison of anomaly detection methods.}
\label{online performance comparison}
\centering
\begin{tabular}{c|c|c|c|c|c|c|c|c|c}
\toprule
Metric & \multicolumn{3}{c|}{I-AUROC} & \multicolumn{3}{c|}{P-AUROC} & \multicolumn{3}{c}{P-AUPRO} \\
\cmidrule(lr){2-4} \cmidrule(lr){5-7} \cmidrule(lr){8-10}
Method & VisA & MVTec AD & MPDD & VisA & MVTec AD & MPDD & VisA & MVTec AD & MPDD \\
\midrule
PatchCore\cite{patchcore} & 0.801 & 0.872 & 0.721 & 0.946 & 0.934 & 0.954 & \textbf{0.938} & \textbf{0.929} & \textbf{0.953} \\
CFA\cite{cfa} & 0.924 & 0.966 & 0.865 & 0.981 & 0.975 & 0.977 & 0.895 & 0.912 & 0.909 \\
FastFlow\cite{fastflow} & 0.883 & 0.927 & \textbf{0.907} & 0.965 & 0.967 & 0.885 & 0.806 & 0.896 & 0.705 \\
DRAEM\cite{draem} & 0.712 & 0.791 & 0.719 & 0.561 & 0.730 & 0.727 & 0.411 & 0.462 & 0.516 \\
Ours-L & \textbf{0.943} & 0.967 & 0.866 & 0.986 & 0.971 & 0.977 & 0.917 & 0.909 & 0.916 \\
Ours-F & 0.934 & \textbf{0.972} & 0.874 & \textbf{0.987} & \textbf{0.976} & \textbf{0.978} & 0.916 & 0.917 & 0.919 \\
\bottomrule
\end{tabular}
\end{table}

\textbf{Computational efficiency.} We also evaluate the computational efficiency of various methods, as shown in Table \ref{inference speed}. The evaluation includes throughput per second (TPS), average time to process one image (TPI), and more specifically, the time for feature extraction and anomaly detection. We conduct five runs for each method, each processing 2000 images from an online input stream, and then calculate the average metrics. Our method demonstrates the highest throughput with 187.117 img/s, and the average anomaly detection time is the lowest at 0.715 ms. This exceptional efficiency can be attributed to the compact size of the memory bank $\mathcal{P}$, facilitating faster computation.

\begin{table}[h]\scriptsize
\caption{Computation efficiency comparison of anomaly detection methods.}
\label{inference speed}
\centering
\begin{tabular}{c|c|c|c|c|c}
\toprule
Metric & PatchCore\cite{patchcore} & CFA\cite{cfa} & FastFlow\cite{fastflow} & DRAEM\cite{draem} & Ours-L/F \\
\midrule
TPS [img/s] & 98.124 & 104.692 & 60.011 & 39.647 & \textbf{187.117} \\
TPI [ms/img]& 10.191 & 9.552 & 16.664 & 25.222 & \textbf{5.344} \\
Encoder [ms] & 6.100 & 5.214 & \textbf{2.302} & 3.591 & 4.629 \\
Detection [ms] & 4.092 & 4.338 & 14.361 & 21.631 & \textbf{0.715} \\
\bottomrule
\end{tabular}
\end{table}

\begin{figure}
  \centering
   \includegraphics[width=\textwidth]{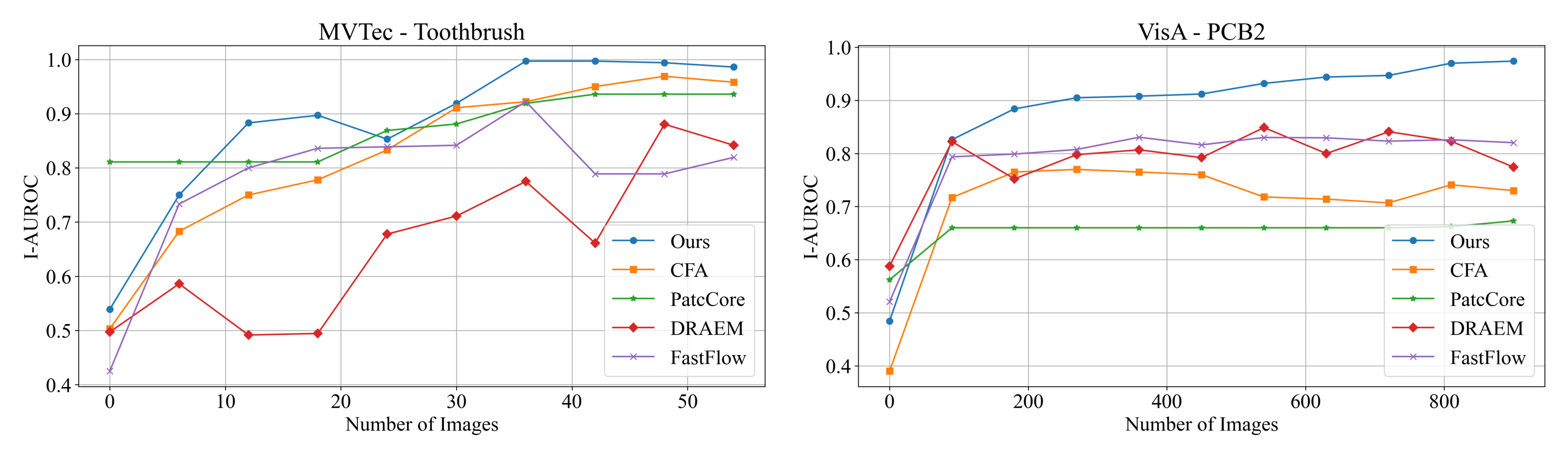}
  \caption{I-AUROC curve for different methods on two example classes. }
  \label{fig:curve_cur}
\end{figure}

\textbf{Convergence curve.} We evaluate which method can achieve higher detection accuracy faster and more stably while receiving an equal amount of online stream data. Each time an image is inputted, the model executes a round of training and anomaly detection. As shown in Fig.\ref{fig:curve_cur}, our method swiftly reaches 90\% I-AUROC accuracy, utilizing approximately 20\% of the data, and ultimately maintains a high level of accuracy. In contrast, PatchCore\cite{patchcore} stops at a lower accuracy after a slow upward. This is because PatchCore's coreset cannot represent the actual distribution of normal samples, since pre-collecting a dataset at the initial stage is not feasible. Additionally, anomaly detection methods such as CFA\cite{cfa} demonstrate good performance in the experiments, mainly due to their ability to adapt image features to new tasks. However, since they cannot obtain a large-scale dataset for establishing a high-quality memory bank in advance, their detection accuracy still leaves room for improvement.

\subsubsection{Robustness for data drift}
\label{exp_sec:Adaptability}
To evaluate the capability of our model to dynamically adapt to data drift in online streaming data, different types of fixed levels of noise are added to the online stream input to simulate changes in image feature distribution. The offline learning scenario with no data drift is the baseline in this experiment. For the data drift scenario, two settings are considered: the offline mode that training with normal samples but testing with drifted data, and the online mode that both training and testing with drifted data. Two types of noise are used to simulate data drift, including fixed-level brightness changes and Gaussian noise. The I-AUROC accuracy report on MVTec AD\cite{mvtec} is shown in Table \ref{tab:noise}.

Although existing methods demonstrate impressive performance in offline settings, where both training and testing samples are well-prepared, they (except for FastFlow\cite{fastflow}) suffer performance degradation when dealing with online data, with a drop in the range of 0.014 to 0.190. This is particularly obvious for reconstruction-based methods such as DRAEM\cite{draem}. Additionally, these methods show poor robustness to cope well with noise. In the offline setting, the most degradation reaches 0.376. It is worth noting that the modifications for online learning can help existing methods handle noises effectively, which proves the strength of online learning in tackling noise. Conversely, our method achieves the best detection accuracy. Furthermore, the online updating mechanism of the memory bank $\mathcal{P}$ enhances the robustness of our method, resulting in only a minor decrease in accuracy by 0.006 and 0.015 under two data drift scenarios.

\begin{table}[h]\scriptsize
\centering
\caption{I-AUROC for different methods under data drift.  Other metrics are in Appendix B.}
\label{tab:noise}
\begin{tabular}{c|c|c|c|c|c|c}
\toprule
Data drift                     & Scenario & Patchcore\cite{patchcore} & CFA\cite{cfa} & FastFlow\cite{fastflow} & DRAEM\cite{draem} & Ours \\ 
\midrule
\multirow{2}{*}{None}           & Offline (B)           & 0.992              & 0.980         & 0.905             & 0.981          & -         \\ 
                                         & Online            & 0.872/\textcolor{gray}{-0.120 }              & 0.966/\textcolor{gray}{-0.014 }          & 0.927 /\textcolor{gray}{0.022 }               & 0.791/\textcolor{gray}{-0.190}             & \textbf{0.967}           \\ 
\midrule
\multirow{2}{*}{Brightness}     & Offline           & 0.914/\textcolor{gray}{-0.078 }     & 0.604/\textcolor{gray}{-0.376} & 0.671/\textcolor{gray}{-0.234}      & 0.692/\textcolor{gray}{-0.289}   & -             \\ 
                                         & Online            & 0.767/\textcolor{gray}{-0.225}       & 0.895/\textcolor{gray}{-0.085} & 0.873/\textcolor{gray}{-0.032}      & 0.695/\textcolor{gray}{-0.286}   & \textbf{0.961}/\textcolor{gray}{-0.006}  \\ 
\midrule
\multirow{2}{*}{Gaussian Noise} & Offline           & 0.722/\textcolor{gray}{-0.270}        & 0.730/\textcolor{gray}{-0.250}   & 0.571/\textcolor{gray}{-0.334}      & 0.680/\textcolor{gray}{-0.301}    & -             \\ 
                                         & Online            & 0.724/\textcolor{gray}{-0.268}       & 0.884/\textcolor{gray}{-0.096} & 0.906/\textcolor{gray}{0.001}      & 0.730/\textcolor{gray}{-0.251}    & \textbf{0.952}/\textcolor{gray}{-0.015}  \\ 
\bottomrule
\end{tabular}
\end{table}

\subsection{Offline learning}
In this section, we display the average I-AUROC accuracy. Performance evaluations of other methods are derived from IM-IAD\cite{im-iad}. As shown in Table \ref{tab:off_comparison}, our method achieves comparable accuracy to the SOTA methods on the full dataset for offline anomaly detection task. Furthermore, few-shot anomaly detection (FSAD) reports demonstrate that our method significantly outperforms these methods by margins of 4.2\%, 4.0\% for the 1, 2-shot cases on the MVTec AD\cite{mvtec} dataset, which means our method have a better performance in scenarios with very limited numbers of samples.

\begin{table}[h]\scriptsize
\caption{I-AUROC comparisons of different methods on full dataset and few-shot scenarios for offline anomaly detection task. Other metrics are in Appendix C.}
\label{tab:off_comparison}
\centering
\begin{tabular}{c|ccc|cccc}
\toprule
 & \multicolumn{3}{c|}{Full Dataset Anomaly Detection} & \multicolumn{4}{c}{Few-shot Anomaly Detection (MVTec AD)} \\
\cmidrule(r){2-4} \cmidrule(r){5-8}
Method & VisA\cite{visa} & MVTec AD\cite{mvtec} & MPDD\cite{mpdd} & 1 & 2 & 4 & 8 \\
\midrule
Patchcore\cite{patchcore} & 0.951 & \textbf{0.992} & \textbf{0.948} & 0.619 & 0.721 & 0.817 & 0.864 \\
CFA\cite{cfa} & 0.920 & 0.980 & 0.923 & 0.813 & 0.839 & 0.879 & \textbf{0.923}\\
FastFlow\cite{fastflow} & 0.822 & 0.905 & 0.887 & 0.552 & 0.552 & 0.729 & 0.801\\
DRAEM\cite{draem} & 0.887 & 0.981 & 0.941 & 0.685 & 0.777 & 0.820 & 0.883\\
Ours & \textbf{0.955} & 0.990 & 0.922 & \textbf{0.855}$\pm$0.0111 & \textbf{0.879}$\pm$0.0130 & \textbf{0.887}$\pm$0.0007 & 0.908$\pm$0.0087 \\
\bottomrule
\end{tabular}
\end{table}

\subsection{Ablations}

We evaluate the impact of memory bank initialization and online update strategies on performance across three datasets. Initialization strategies considered: (1) random single image, (2) undecoupled random noise, and (3) decoupled (QR decomposition) random noise. Update strategies included: (1) updating image features without changing the memory bank, (2) learning-based update (Section \ref{learning based strategy}), and (3) feature-enhanced update (Section \ref{feature based strategy}). I-AUROC accuracies are reported in Table \ref{ablation 2}. More details could be found in Appendix D.

\begin{table}[ht]\scriptsize
\caption{Impact of memory bank initialization and update strategies. The memory bank update strategies are without update (w/o), learning-based update (L), and feature-enhanced update (F).}
\label{ablation 2}
\centering
\begin{tabular}{l|ccc|ccc|ccc}
\toprule
& \multicolumn{3}{c|}{Random single image} & \multicolumn{3}{c|}{Random noise} & \multicolumn{3}{c}{Decoupled random noise} \\
\cmidrule(lr){2-4} \cmidrule(lr){5-7} \cmidrule(lr){8-10}
Dataset  & w/o & L & F & w/o & L & F & w/o & L & F \\
\midrule
MVTec AD\cite{mvtec} & 0.967 & 0.969 & 0.974 & 0.778 & 0.788 & 0.738 & 0.967 & 0.967 & 0.972 \\
MPDD\cite{mpdd} & 0.859 & 0.876 & 0.874 & 0.627 & 0.627 & 0.734 & 0.862 & 0.866 & 0.874 \\
VisA\cite{visa} & 0.919 & 0.938 & 0.927 & 0.706 & 0.743 & 0.688 & 0.936 & 0.943 & 0.934 \\
\bottomrule
\end{tabular}
\end{table}

\textbf{Memory bank initiation.} When using the same update strategy, initialization via a random single image and decoupled random noise outperform undecoupled random noise. On the MVTec AD\cite{mvtec} dataset, the corresponding I-AUROC accuracies are 97.4\%, 97.2\% compared to 73.8\%, on the MPDD\cite{mpdd} dataset, they are 87.4\%, 87.4\% compared to 73.4\%, and on the VisA\cite{visa} dataset, they are 92.7\%, 93.4\% compared to 68.8\%. In a word, the strategy of initializing the memory bank based on decoupling random noise is efficient, yielding a detection accuracy on par with the random image initialization strategy. The learning-to-update process can incorporates the discriminative normal pattern into prototype memory bank, while decoupling plays a crucial role to ensure the implementation of this process firmly.

\textbf{Memory bank update strategy. } With the same initialization, their I-AUROC metrics progressively increase from without, learning-based to feature-enhanced updates. Interestingly, under the initialization setting of random noise without decoupling, the feature-enhanced memory bank update strategy improves the I-AUROC metric by 10.7\% on MPDD\cite{mpdd} compared to the first two strategies. This suggests that under poor initialization conditions, the feature-enhanced update strategy can provide more constraints to ensure the detection lower bound. However, due to its computational cost, we recommend the learning-based update for balancing efficiency and accuracy.

\textbf{Memory bank size.} The impact of memory bank size $K$ on detection accuracy is also evaluated (see Appendix D for details). Our findings indicate that when the features stored in the memory bank exhibit desirable decoupling properties, even a smaller set of features can achieve accuracy comparable to that of SOTA methods. Our method, therefore, significantly diminishes memory bank size compared to previous methods, enhancing online anomaly detection. More broadly, the orthogonal decoupling-based memory bank filling method proposed in this paper can potentially function as a generalized principle, benefiting various embedding-based anomaly detection methods. 

\section{Conclusion}
%在本文中我们第一个关注了在线图像异常检测算法，它在工业场景尤其是流程制造场景下更具现实的、重要的应用价值。我们深入细致地分析了现有基于deep feature embedding方法在online设置下的不足，并创新性地提出了正交化的memory初始化方法，避免了大量样本的准备过程。除此之外，我们提出了一个简单但有效的基于类内对比学习方法，实现了原型memory bank和偏置特征的联合更新。这些使得我们方法可以实现完全在线的学习，大量的实验结果证明了所提出方法在在线和离线环境下都取得了优异的性能表现。

% In this paper, we focus on online image anomaly detection algorithms, which have more realistic and important applications in industrial scenarios, especially in process manufacturing scenarios. We analyze in detail the shortcomings of existing deep feature embedding-based methods and innovatively propose an orthogonalized memory initialization method to avoid the preparation process of a large number of samples. In addition, we propose a simple but effective intra-class comparison-based learning method that implements the joint update of the prototype memory bank and biased features. This allows our method to achieve fully online learning, and extensive experimental results demonstrate that the proposed method achieves excellent performance in both online and offline environments.

In this paper, we propose an online image anomaly detection method, namely \textit{LeMO}, which carries significant and realistic application value in industrial scenarios, especially in process manufacturing scenarios. We conduct an in-depth and detailed analysis of the limitations of previous deep feature embedding methods in the online setting, and propose utilizing an orthogonal memory initialization method to obviate the need for the cumbersome pre-collection of a large number of samples. Furthermore, we propose a simple and effective intra-class contrast learning based anomaly detection method, which enables the online joint optimization of prototype memory bank and biased features. LeMO achieves fully online anomaly detection and localization. Extensive experimental results demonstrate superior performance of LeMO in both online and offline environments.

%Discussion 在实践中发现，使用预训练好的backbone去提取特征，不仅仅可能会导致特征偏置需要特征更新以抵消，还是阻碍模型进一步提升运行效率的关键因素。本文中，受到时间和精力限制，我们直接使用了预训练的模型，但更加轻量化、更加适配工业场景的backbone是非常有意义的研究路径。除此之外，虽然本文方法训练和测试均十分高效，但其仍有部分参数需要训练，我们未来可考虑无参数不需要训练的策略。

% \textbf{Discussion.} In practice, it is found that using a pre-trained backbone may not only lead to mismatch feature with bias, but is also a key factor that hinders the total model's operational efficiency. In this paper, we directly use the pre-trained model. A more lightweight backbone with higher adaptability for industrial vision is the meaningful pathway. In addition, although the method in this paper is very efficient in both training and inference, some of its parameters still need to be trained, we will explore the strategy without any trainable parameters in the future.

\textbf{Discussion.} In practice, we have observed that using a pre-trained backbone may not only lead to feature bias, but also hinders the model's further efficiency improvement. In this paper, due to time constraints, we directly use a pre-trained model. However, research into a more lightweight backbone that is better adapted to industrial scenarios is a significant and meaningful path. Furthermore, although our method is highly efficient in both training and testing, there are still some parameters that require training. In the future, we may consider strategies that eliminate the need for any parameter training, aiming to conduct more efficient anomaly detection.

% \section*{References}
\bibliographystyle{plain}
\bibliography{main}

%%%%%%%%%%%%%%%%%%%%%%%%%%%%%%%%%%%%%%%%%%%%%%%%%%%%%%%%%%%%

\end{document}